\documentclass[runningheads]{llncs}

\usepackage[mobile]{eccv}

\usepackage{eccvabbrv}

\usepackage{graphicx}
\usepackage{booktabs}

\usepackage[accsupp]{axessibility}  

\usepackage{hyperref}

\usepackage{orcidlink}
\usepackage[dvipsnames]{xcolor}
\definecolor{blue_mod}{RGB}{0,32,96}
\definecolor{red_mod}{RGB}{192,0,0}

\usepackage{breqn}
\usepackage{amsmath,bm}
\usepackage{algpseudocode}

\DeclareMathOperator*{\argmin}{argmin}   
\usepackage{enumitem}
\DeclareMathOperator*{\idx}{idx}
\usepackage[dvipsnames]{xcolor}
\usepackage{gensymb}
\usepackage[]{multirow}
\usepackage{graphbox}
\usepackage{algorithm,stfloats}
\newcommand{\rone}[1]{\textcolor{red}{R1}} 
\newcommand{\rtwo}[1]{\textcolor{green}{R2}} 
\newcommand{\rthree}[1]{\textcolor{blue}{R3}} 

\begin{document}

\title{COSMU: Complete 3D human shape from monocular unconstrained images}


\author{Marco Pesavento\orcidlink{0000-0003-2379-6593} \and Marco Volino\orcidlink{0000-0002-1869-9257} \and
Adrian Hilton\orcidlink{0000-0003-4223-238X}}

\authorrunning{M.~Pesavento et al.}

\institute{Centre for Vision, Speech and Signal Processing (CVSSP), University of Surrey, Guildford, UK
\email{\{m.pesavento,m.volino,a.hilton
\}@surrey.ac.uk}}

\maketitle

\begin{abstract}
We present a novel framework to reconstruct complete 3D human shapes from a given target image by leveraging monocular unconstrained images. The objective of this work is to reproduce high-quality details in regions of the reconstructed human body that are not visible in the input target. The proposed methodology addresses the limitations of existing approaches for reconstructing 3D human shapes from a single image, which cannot reproduce shape details in occluded body regions. The missing information of the monocular input can be recovered by using multiple views captured from multiple cameras. However, multi-view reconstruction methods necessitate accurately calibrated and registered images, which can be challenging to obtain in real-world scenarios.
Given a target RGB image and a collection of multiple uncalibrated and unregistered images of the same individual, acquired using a single camera, we propose a novel framework to generate complete 3D human shapes. We introduce a novel module to generate 2D multi-view normal maps of the person registered with the target input image. The module consists of body part-based reference selection and body part-based registration. The generated 2D normal maps are then processed by a multi-view attention-based neural implicit model that estimates an implicit representation of the 3D shape, ensuring the reproduction of details in both observed and occluded regions.
Extensive experiments demonstrate that the proposed approach estimates higher quality details in the non-visible regions of the 3D clothed human shapes compared to related methods, without using parametric models.
\end{abstract}
    
\section{Introduction}
\label{sec:1_intro}
Democratizing the creation of high-quality human avatars is essential to unlock a wide range of immersive technologies including self-representation in the metaverse, virtual conferencing, and virtual try-on. Realistic human avatars are crucial in these applications, making 3D human reconstruction a prominent topic in computer vision and graphics. Research focuses on finding practical solutions to replace existing sophisticated multi-view systems for 3D reconstruction, which are inaccessible to the general community. Single-view 3D human reconstruction methods have been developed to reconstruct 3D human shapes using a simple and cost-effective set-up, such as a single camera. 
\\Recent methods~\cite{saito2019pifu,he2020geopifu,alldieck2022photorealistic,corona2023structured,he2021arch++,huang2020arch} use neural implicit models to estimate 3D human shapes with high-quality details in visible regions from single RGB images. These works fail to reproduce fine detail in non-visible regions of the input image, resulting in overly smooth reconstruction.
\\To address this problem, solutions extend single-view to multi-view reconstruction, using multiple images of various orientations of the human~\cite{yu2023multi,li2022detailed,zins2021data,zins2023multi,cao2023sesdf}. Multi-view methods represent details across the entire body but require calibrated cameras and registered images, with the subject in the same pose and clothing. Systems to capture such images are inaccessible to the general community in real-world scenarios, significantly limiting reproducibility.
\begin{figure}[t]
\centering
\includegraphics[width=0.95\linewidth]{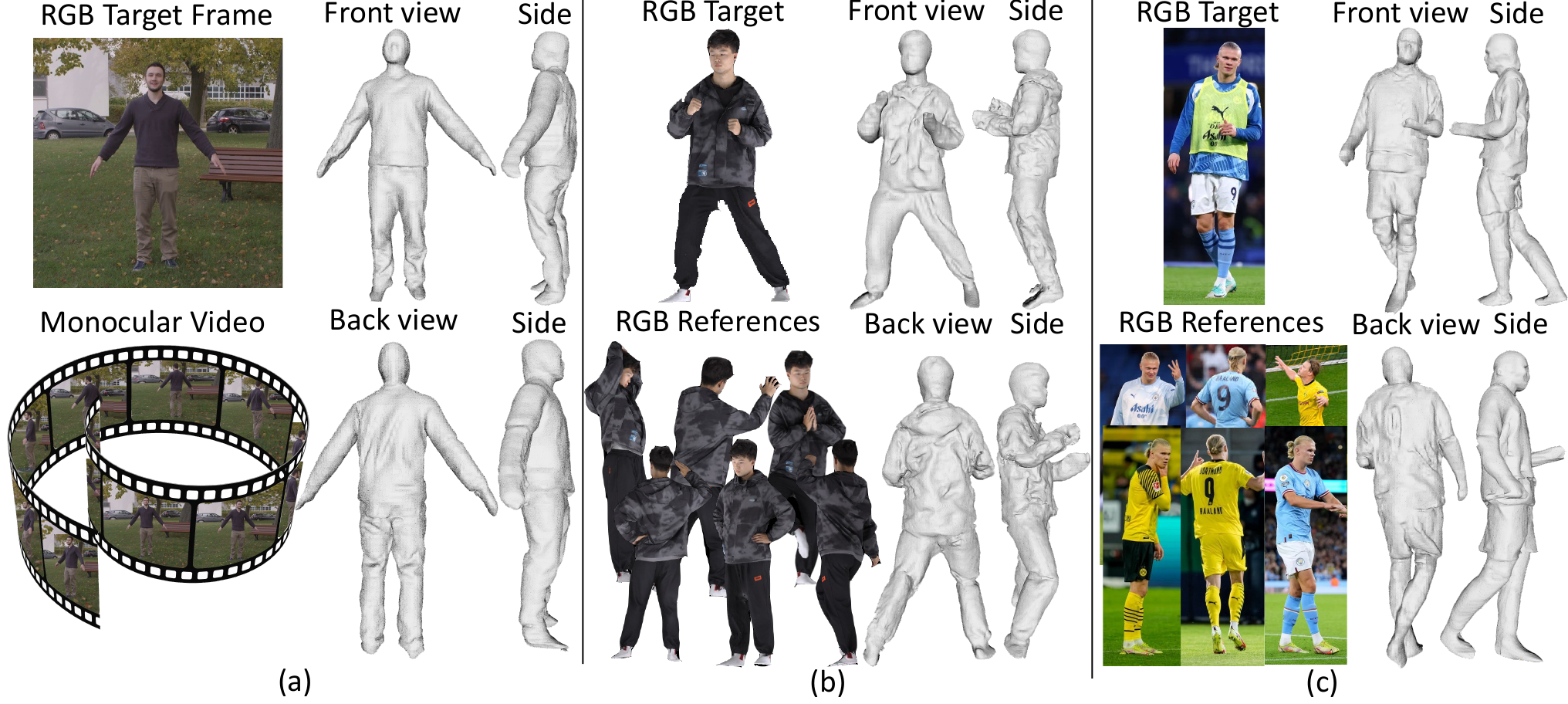}
\caption{COSMU reconstructs complete 3D human shape from a target RGB input by leveraging a set of monocular images collected from various scenarios, including monocular video (a), images with different poses (b) and different clothing (c).}
\label{fig:introduction}
\end{figure}
%
%
\\Recent works~\cite{zhu2022mvp, yu2022multiview,huang2022unconfuse,zheng2021pamir,chen2022crosshuman} reconstruct humans from a single target image by leveraging multiple monocular unconstrained images. We define images as unconstrained if camera calibration parameters are not known (uncalibrated) and the images are captured at different time instances resulting in pose and clothing variations without spatial alignment (unregistered). These methods directly fuse features from multiple images, which is a challenging task since diverse views and poses must be fused into a single coherent reconstruction~\cite{zhu2022mvp}. They also rely on parametric models and cannot use images of the person in different clothing.
\\We propose COSMU, a novel framework that reconstructs COmplete 3D human Shape from Monocular Unconstrained images. Our objective is to reconstruct a 3D clothed human shape from a single target image, focusing on reproducing high-quality details in regions of the reconstructed human body that are not visible in the input target. To achieve this, we leverage a collection of monocular unconstrained images of the same individual. COSMU does not rely on parametric models in the reconstruction process as related works.
\\The multi-view scenario used by the multi-view reconstruction approaches is simulated from a collection of unconstrained images, with the subject in different poses and clothing and without camera calibration or correspondences. 
We present a novel pipeline that applies an optimization-based algorithm to select, from the set of unconstrained images, $N_{r}$ references for each $N_{p}$ body part of the target image following an orientation-visibility-similarity criteria. 
%
A neural implicit model generates a coarse point cloud $\zeta_t$ of the target and of each selected reference $\zeta_{ref}$. The 3D points of these point clouds are classified into $N_{p}$ body parts using a novel pixel-aligned neural network. The corresponding reference body part selected for a specific target part is registered to it. To obtain $N_r + 1 $ views of the subject, 2D normal maps are generated from the registered point clouds.  
The multi-view scenario is simulated and the 2D normal maps are processed by an attention-based implicit model that estimates the implicit function $f_f$ of the complete 3D human shape. 
COSMU generates 2D multi-view normal maps instead of RGB images because the references are not constrained to the same clothing as the target and RGB value estimation of 3D points is avoided.
\\
Extensive experiments show COSMU's ability to reconstruct complete 3D human shapes from monocular images taken from various scenarios, including monocular videos (\cref{fig:introduction}a), images with different poses and same clothing (\cref{fig:introduction}b)  and images with different poses and clothing (\cref{fig:introduction}c). Thanks to the flexibility in creating the reference set, COSMU represents a significant advance for high-quality 3D avatar reconstruction.
\\The contributions presented in this paper include:
\begin{itemize}[noitemsep]
    \item A novel framework to reconstruct complete 3D human shapes from a target RGB image by leveraging a set of monocular uncalibrated and unregistered images without the use of parametric models.
    \item A novel pipeline to simulate a multi-view scenario from a collection of unconstrained reference images via reference selection and body part registration.
    \item Improved performance for human shape reconstruction compared to state-of-the-art methods with higher quality details in the regions of the reconstructed human shape that are occluded in the target image.
\end{itemize}

\section{Related Works}
\label{sec:2_related}
\textbf{3D human reconstruction from monocular cameras}
Early works on 3D human reconstruction from monocular cameras represent the body shape without capturing shape details~\cite{bogo2016keep,kanazawa2018end,pavlakos2019expressive,kocabas2020vibe,xu20213d,varol2018bodynet,zheng2019deephuman,alldieck2018video,alldieck2019tex2shape,alldieck2019learning}. The introduction of neural implicit models~\cite{mescheder2019occupancy,park2019deepsdf,chen2019learning} revolutionized 3D modeling by estimating an implicit representation, which reproduces higher-quality shapes due to its inherent property of defining a surface as a level set of an occupancy probability function. PiFU~\cite{saito2019pifu} introduced pixel-aligned implicit function for human reconstruction, aligning 3D points with features of the input image to achieve an unprecedented level of detail in the reconstructed shapes. 
Since then, neural implicit models have become the standard representation for 3D human shapes~\cite{he2020geopifu,huang2020arch,alldieck2022photorealistic,liao2023high,corona2023structured}.
\\A significant limitation of these approaches is their inability to reproduce high-quality details in non-visible regions. 
SuRS~\cite{SuRSECCV2022} enhances shape quality by learning a map from a low to a high-resolution shape, hallucinating details in occluded regions, which are still overly smoothed as no information is provided to the implicit model. 
To improve the quality of non-visible regions, PiFU-HD~\cite{saito2020pifuhd} generates front and back surface normals of the human from the RGB image. This estimation introduces noise, resulting in smoothed surfaces of occluded regions. 
ICON~\cite{ICONCVPR2022} improves normal estimation by guiding it with SMPL~\cite{loper2015smpl} fitted to the image. Similarly, ECON~\cite{xiu2023econ} alleviates the SMPL problem of clothing reconstruction by feeding the estimated front and back normals into a d-BiNI optimizer.
Conditioning the learning process on SMPL body poses a challenge for the network, which cannot fully exploit 2D image features. These methods estimate only an additional view, which is insufficient to represent the entire body and its accuracy influences the reconstruction quality of the non-visible regions.
TeCH~\cite{huang2023tech} proposes to estimate the unseen view from the input through Score Distillation Sampling~\cite{poole2022dreamfusion} (SDS) and represents the human with drivable tetrahedral representation (DMTet~\cite{shen2021deep}). Subject-tailored fine-tuning is required, limiting its generalizability. Other works focus only on 3D reconstruction from monocular videos, requiring to optimize their algorithms on a single monocular video, limiting the generalization and their application scenarios~\cite{jiang2022selfrecon,alldieck2019learning,alldieck2018video,alldieck2018detailed}.   NERF-based methods are applied to 3D human reconstruction from sparse views, but their subject-specific nature limits their utility~\cite{peng2020neural,chen2021animatable,te2022neural,jayasundara2023flexnerf,li2023ghunerf,guo2023vid2avatar,wang2022arah}. For a fair comparison with the works that require subject-specific optimization, our approach should be fine-tuned with the data of each specific subject, significantly limiting generalizability. Therefore, these works are not evaluated in this paper. 
\\\textbf{3D human shape from unconstrained RGB images}
\\PaMIR~\cite{zheng2021pamir} extended single-view reconstruction to multiple unconstrained images by using the SMPL model as a body reference to establish correspondences across different images. Latent embedding vectors from the correspondent frames are aggregated into a single embedding using mean pooling and then processed by a multi-layer perception (MLP) to estimate the implicit representation. 
Similarly, MVP-Human\cite{zhu2022mvp} fits the SMPL model to each frame to establish correspondences between unconstrained images. Skinning weights of a canonical pose are estimated to guide the fusion of the features from multiple images through an attention-based implicit model.
%
%
UnconFuse~\cite{huang2022unconfuse} builds upon MVP-Human, with a focus on estimating  T-pose shapes from unconstrained images, while posed shapes cannot be reconstructed. 
CrossHuman~\cite{chen2022crosshuman} selects references following PoseFusion~\cite{li2021posefusion} and leveraging the SMPL model. Features are then extracted from both the fitted SMPL and RGB input and fused with a transformer to estimate the implicit representation.
These works extract features from unconstrained images and fuse them with a neural model, losing details in regions where the body poses differ among frames. In contrast, COSMU generates 2D multi-view normals of the human in a spatially aligned pose, facilitating the estimation of the final implicit function. Moreover, these works rely on parametric models, whose fitting can introduce noise. COSMU distinguishes itself by not relying on parametric models.
\\ Have-Fun~\cite{yang2023have} estimates a DMTet~\cite{shen2021deep} representation by integrating SDS~\cite{poole2022dreamfusion} and SMPL skinning weights to adapt the unconstrained images to the target one. Similarly to Tech~\cite{huang2023tech}, fine-tuning on the target subject is required, limiting generalizability. Other works focus on 3D human reconstruction from other unconstrained data such as depth images~\cite{li2021posefusion, pesavento2024anim} or 3D meshes~\cite{li2022avatarcap}, which is beyond the scope of this paper.

\begin{figure*}[t!]
\centering
\includegraphics[width=\linewidth]{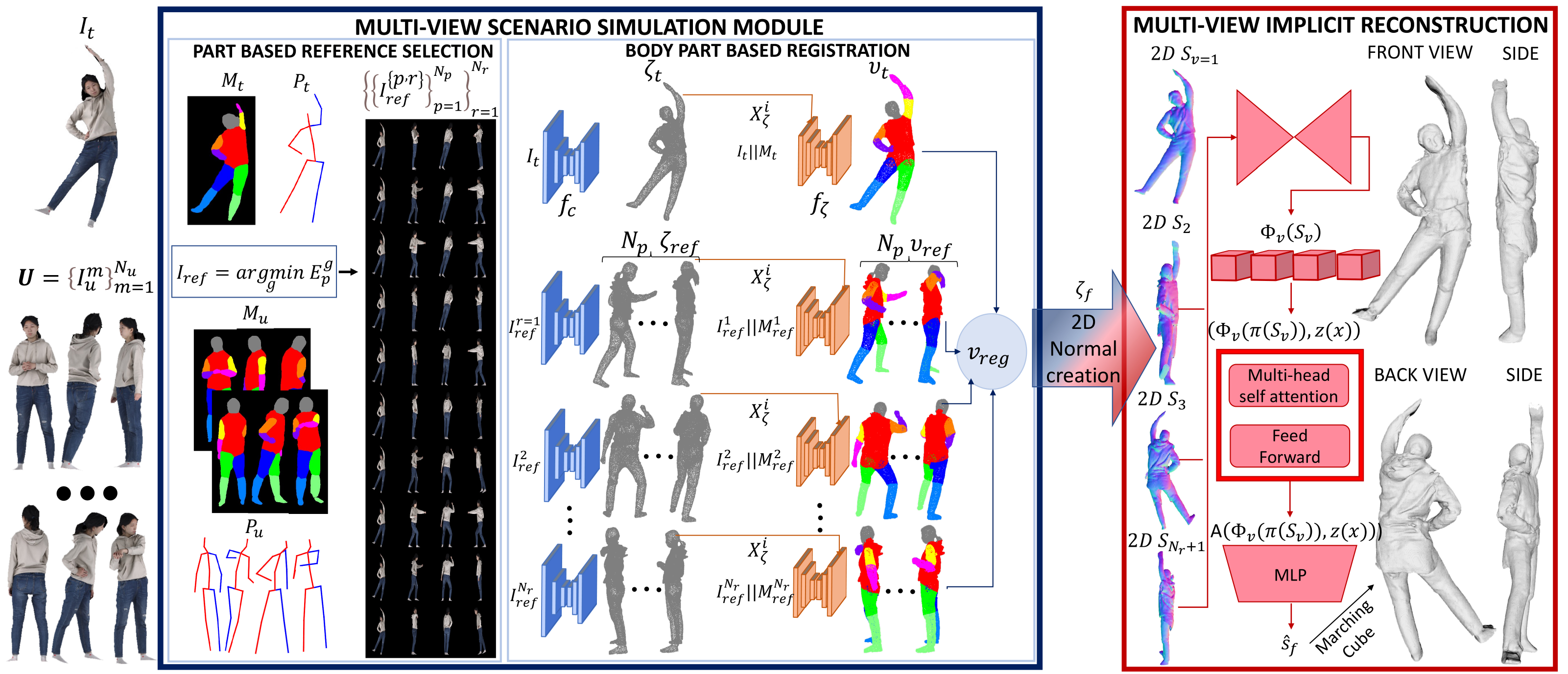}
\caption{COSMU framework: given a single RGB image as input and a collection of monocular unconstrained images, a {\color{blue_mod}multi-view scenario simulation module}, which consists of part-based reference selection and registration, is first applied to generate registered multi-view 2D normal maps of the subject in the same pose. An implicit function $\hat{s}_f$ is then estimated by processing these normal maps with a {\color{red_mod}multi-view attention-based neural implicit model}. The 3D mesh is reconstructed via marching cube.}
\label{fig:arch}
\end{figure*}
\section{3D human shape reconstruction from monocular unconstrained images}
\label{sec:3_methods}
We introduce a novel framework that reconstructs COmplete 3D human Shape from Monocular Unconstrained images (COSMU). To be complete, a 3D human shape should contain high-quality details in every body region. A single image lacks the necessary information to reproduce details across the entire body, which can instead be extracted from multi-view images. In contrast to multi-view reconstruction, which requires calibrated and registered images captured with inaccessible systems, COSMU, given a single target image, reproduces fine details in the regions of the 3D shape that are occluded in the input image. This is achieved by leveraging a collection of uncalibrated and unregistered monocular images, referred to as unconstrained images. The input is a target image $I_t$ of a person together with multiple views of the same person in various poses and either same or different clothing, from which we extract body part detail non-visible in the target view. As shown in~\cref{fig:arch}, given a set $\mathbf{U}=\{I^m_u\}^{N_u}_{m=1}$ of $N_u$ monocular unconstrained images, COSMU simulates the multi-view scenario of traditional multi-view works using a novel module that employs body part-based reference selection and registration to create 2D multi-view normal maps registered with the target input. A multi-view attention-based implicit model regresses the implicit representation $f_{f}$ of the complete 3D shape from the normals.
\subsection{Multi-view scenario simulation module}
Directly processing unconstrained images with a multi-view neural model deteriorates the performance due to their unregistered nature (\cref{ssec:ablation}). To address this, we propose a novel pipeline that simulates a multi-view scenario in which the multi-view images represent various orientations of the subject in the same body pose. Given a target image $I_{t}$, we select $N_{p} \times N_{r}$ references $\left\{\{I_{ref}^{\{p,r\}}\}^{N_p}_{p=1}\right\}^{N_r}_{r=1}, I_{ref} \in \mathbf{U}$, from the collection $\mathbf{U}$ of unconstrained images applying a body part-based reference selection. 2D registered multi-view normal maps are then created by registering references to the target input via body part-based registration. Set notation is dropped for simplicity.
\\\textbf{Body part-based reference selection: }To improve the body part-based registration, we select references observed from complementary orientations to the target one and with body part poses similar to the corresponding target body part ones. $N_r$ references are selected for each $N_p$ body part of the target $I_t$ via energy optimization following an orientation-visibility-similarity criteria.  Semantic masks $M$ and 3D poses $P$ are extracted from the input images. To ensure that the entire body is represented in the selected references, we first divide the $N_u$ unconstrained images into $N_r$ groups based on the full-body orientation angle $\theta_u$ computed between the reference $P_u$ and the target $P_t$ 3D poses with the Rodrigues rotation formula. Each group contains a variable number $N_g$ of unconstrained images $I^g_u, g \in [1,\ldots,N_g]$. $N_p$ references $I_{ref}$ are selected from each group following 3 criteria:
\\\textbf{1.} Body orientation angle: for a complete view of the body, the selected images should complement each other in the body orientation. An orientation metric is introduced to enforce the selection of images in which the body orientation is closer to the central angle $\theta^r_c$ of the $r^{th}$ group:
\begin{equation}
\label{eq:orientation}
    o_g=|\theta^r_c-\theta_u|,\quad \theta^r_c=r\cdot\frac{360}{N_r+1},\quad r\in[1,...,N_r]
\end{equation}
\\\textbf{2.} Body part visibility: every body part should be visible in the selected references to have a complete view of the body. We use a visibility metric that counts the total number of pixels $N_M$ of each body part $p$ in the semantic masks $M$.
\begin{equation}
\label{eq:visibility}
    v_g^p=N^p_M,\quad p\in[1,...,N_p]
\end{equation}
\\\textbf{3.} Body part pose similarity: similarity between each body part of $I^g_u$ and the corresponding body part of $I_t$ is evaluated as 3D Euclidean distance between the corresponding 3D bones of $P_g$ and $P_t$, ensuring spatial consistency:
\begin{equation}
\label{eq:similarity}
    b_g^p=||P^p_t-P^p_g||,\quad p\in[1,...,N_p]
\end{equation}
$o_g=o(I^g_u), v^p_g=v^p(I^g_u), b^p_g=b^p(I^g_u)$ for the $g^{th}$ image $I^g_u$ of the $p^{th}$ body part.
\\$N_p$ references $I_{ref}$ are selected by minimizing the energy $E^p_{g}$ over the $N_g$ images $I^g_u$ for each body part $p$:
\begin{equation}
\label{eq:energy}
\begin{gathered}
   I_{ref}=\argmin_{g}(E^{p}_{g})\\
  \resizebox{0.65\hsize}{!}{$  E^{p}_{g}=idx_{min}\left(o_g\right)+idx_{max}\left(v^p_g\right)+idx_{min}\left(b^p_g\right)$}
\end{gathered}
\end{equation}
$idx_{min}$ indicates the position $idx$ of the $g^{th}$ image $I^g_u$ in the arrays created by sorting all the $N_g$ images with respect to their specific metric $K=o_g, v_g^p, b_g^p$. $idx_{max}$ sorts the array in inverse order:
\begin{equation}
\begin{gathered}
   \resizebox{0.75\hsize}{!}{$
   idx_{min}(K)= \idx_{I^g_u} \{I^1_u,\ldots,I^{N_g}_u\}, \resizebox{0.28\hsize}{!}{${K(I^1_u) \leqslant \ldots \leqslant K(I^{N_g}_u)}$}$}\\
  \resizebox{0.75\hsize}{!}{$ idx_{max}(K)= \idx_{I^g_u} \{I^1_u,\ldots,I^{N_g}_u\}, \resizebox{0.28\hsize}{!}{${K(I^1_u) \geqslant \ldots \geqslant K(I^{N_g}_u)}$}$}
\end{gathered}
\end{equation}
$N_p \times N_r$ references are selected by repeating the body part-based reference selection for every group.
\\\textbf{Body part-based registration: }
To simulate the multi-view scenario given the selected $N_p \times N_r$ references, we register each 3D reference body part to the corresponding 3D body part in $I_t$. This process begins with the reconstruction of a coarse point cloud $\zeta_t$ for $I_t$ and for each of the $N_p$ reference images $I_{ref}$ previously selected for the $N_r$ groups ($\zeta_{ref}$). We use a neural implicit model $f_{c}$ comprising of a U-Net that increases the resolution of the input features $\phi(I)$ as in SuRS~\cite{SuRSECCV2022}. The coarse implicit function is estimated as:
\begin{equation}
\label{eq:coarse_point}
\hat{s}_c=f_{c}\Bigl(\phi\bigl(\pi(I)\bigr),z(x)\Bigr), \quad \hat{s}_{c}\in \mathbb{R}
\end{equation}
where $I$ is either $I_t$ or $I_{ref}$, $\pi$ represents the orthographic projection of 3D points $x$ into the input images $I$ and $z(x)$ is the depth of $x$. $f_c$ is trained with:
\begin{equation}
    \mathcal{L}_{c}=\frac{1}{N}\sum_{i=1}^{N}\left|f_{c}(X^i_{c})-f_{c}^{gt}(X^i_{c})\right|^2 + \left|I-I_{HR}\right|
\end{equation}
where $X^i_{c}=(\phi^i(\pi(I)),z(x))$, $N$ is the number of points in space and $I_{HR}$ is the up-sampled ground-truth version of $I$ used to train the U-Net feature extractor.
\\We then retrieve 3D body parts by classifying each 3D point $x_{\zeta}$ of the point clouds $\zeta_{t/ref}$ with a novel pixel-aligned classification network $f_{\zeta}$. An image feature embedding $\phi(I||M)$ is extracted from the concatenation between the input RGB image and its semantic mask $M$. Each $x_{\zeta}$ is projected using orthographic projection $\pi$ onto the input $I||M$ and indexed to $\phi(I||M)$. An MLP processes this embedding to estimate the body part $\hat{s}_{\zeta}$ to which $x_{\zeta}$ belongs:
\begin{equation}
    \label{eq:part_class}
    \hat{s}_{\zeta}=f_{\zeta}\left(\phi(\pi(I||M)),z(x_{\zeta})\right),\quad \hat{s}_{\zeta}\in [1,...,N_p]
\end{equation}
$N_p$ 3D body parts $\upsilon_{ref}$ are defined for each reference point cloud as well as for the target $\upsilon_t$. Next, the specific $p^{th}$ body part $\upsilon_{ref}$ selected for the corresponding $p^{th}$ target body part $\upsilon_t$ is registered with it as $\upsilon_{reg}=R(\upsilon_{ref})+ T $,
where $R$ is the rotation and $T$ the translation computed via rigid registration. For each group, the registered $N_p$ body parts are combined to create a point cloud $\zeta_{f}$. $N_r$ point clouds $\zeta_{f}$ registered with $\zeta_t$ are generated, rotated by their $\theta_u$ and, together with $\zeta_t$, used to retrieve $N_r +1$ 2D normal maps $S_v$, one for each view $v \in [1,\ldots,N_r+1]$. Normal maps are generated instead of RGB images to avoid the estimation of color values for each 3D point of $\zeta_{t/ref}$ and to allow different clothing in the references. Directly fusing the registered point clouds to obtain the 3D shape is challenging since correspondences are unknown. 
We train $f_{\zeta}$ with a Mean Squared Error $MSE$ loss:
\begin{equation}
    \mathcal{L}_{part}=\frac{1}{N}\sum_{i=1}^{N}\left|f_{\zeta}(X^i_{\zeta})-f_{\zeta}^{gt}(X^i_{\zeta})\right|^2
\end{equation}
where $X^i_{\zeta}=\Bigl(\phi^i\bigl(\pi(I||M)\bigr),z(x_{\zeta})\Bigr)$
\subsection{Multi-view implicit representation}
Neural implicit models have proven to be effective in reproducing high-quality details in reconstructed 3D shapes. This is attributed to the characteristic of the implicit function that does not discretize the 3D space like other representations~\cite{saito2019pifu}. Leveraging multiple views to learn the implicit representation further enhances the quality of the reconstruction, as it enables the incorporation of information about the entire body. Concatenating features from multiple images is a suboptimal approach because it imposes an order on the views~\cite{zins2021data}. Fusion approaches based on statistics have been proposed~\cite{gardner2019classifying,su2015multi} but their pooling operation causes a loss of information regarding the individual view contributions. Inspired by Zins \etal~\cite{zins2021data}, we propose to fuse the features extracted from the multi-view 2D normals using an attention-based architecture. An attention score is computed for each view $v$ based on the compatibility of a query with a corresponding key, resulting in the generation of $N_{r}+1$ features. Each feature contains the original information from the corresponding view, merged with the information from all the other views. An MLP then models a pixel-aligned implicit function $f_f$ to learn the occupancy value  $\hat{s}_f$ of the 3D points in space:
\begin{equation}
\label{eq:final_estimation}
\hat{s}_f=f_f\left(A\left(\phi_v(\pi(S_v)),z(x)\right)\right),\quad \hat{s}_f \in \mathbb{R}
\end{equation}
where $\phi_v$ are the feature extracted from each view $v$ of the surface normal $S_v$, $\pi$ indicates the pixel-alignment between the  3D points of the space and each $\phi_v$, $A$ is the attention operation and $z(x)$ is the depth value of the 3D points $x$.  We train the multi-view architecture with an MSE loss:
\begin{equation}
    \mathcal{L}_{f}=\frac{1}{N}\sum_{i=1}^{N}\left|f_{f}(X^i_{f})-f_{f}^{gt}(X^i_{f})\right|^2
\end{equation}
where $X^i_{f}=A\left(\phi_v^i(\pi(S_v)),z(x^i)\right)$.
\\\textbf{Inference:} $N_p \times N_r$ references are selected from the collection of unconstrained images. $N_r+ 1$ 2D normal maps are generated using body part-based registration to simulate a multi-view scenario with registered views of the subject. The normals are processed by the multi-view attention-based model to estimate the implicit representation. The 3D mesh is reconstructed via marching cube~\cite{lorensen1987marching}.
\\To achieve pixel-alignment through projection, multi-view works require the parameters of calibrated cameras in specialized capture systems. In COSMU, instead of using the camera projection matrix for pixel-alignment, 3D points of the space are rotated by $\theta_u$ computed by the multi-view scenario simulation module and projected onto the images using an identity matrix. 
Existing approaches create the inference 3D grid centered at the origin. If the subject is not centered in the input images, the multi-view pixel-alignment of COSMU is inaccurate since the same rotated 3D point may not be projected in the corresponding pixel in each view. To tackle this, we set the boundary limits of the 3D grid as the boundaries of the coarse point cloud $\zeta_t$, ensuring multi-view pixel-alignment.
\section{Experiments}
\label{sec:4_experiments}
\begin{figure*}[t!]
\centering
\includegraphics[width=0.95\linewidth]{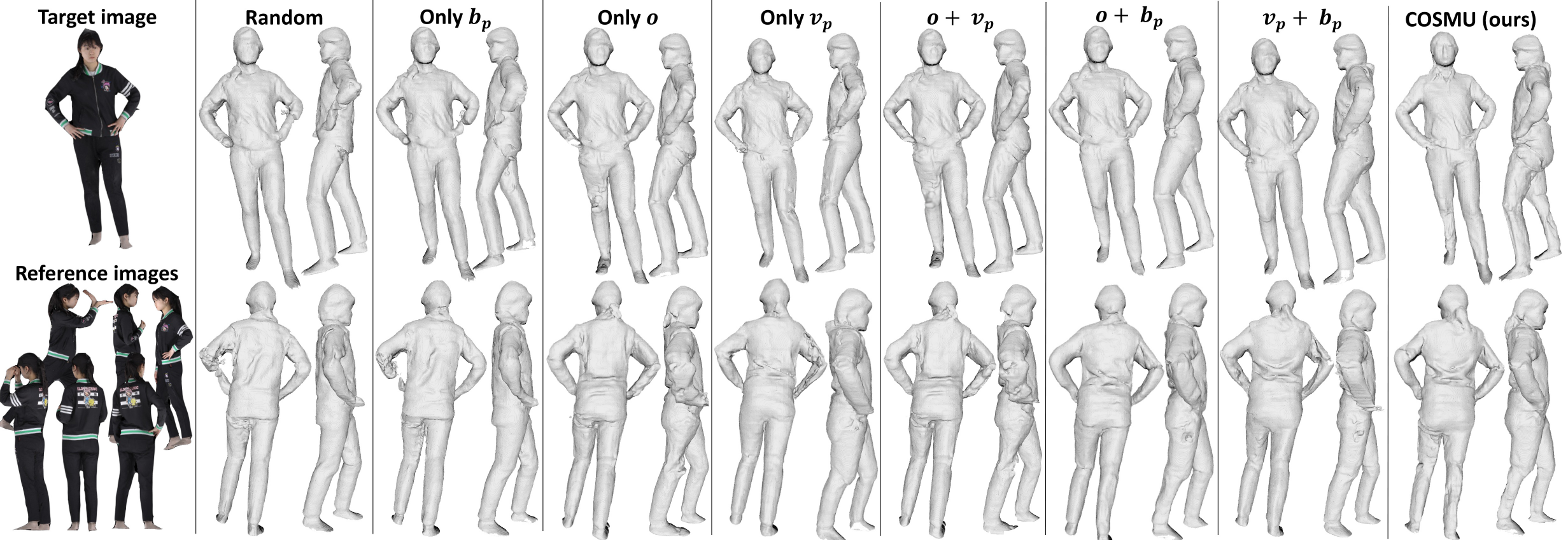}
\caption{Qualitative comparisons between 3D shapes from THuman3.0~\cite{deepcloth_su2022} reconstructed by changing the energy optimization process to select reference images. Front, back and side views of the shapes are illustrated.}
\label{fig:ablation_energy}
\end{figure*}
We conduct a comprehensive evaluation of COSMU for the task of 3D human shape reconstruction from a target image by leveraging a collection of monocular unconstrained images of the same individual. The neural models of the framework are trained with THuman2.0 dataset~\cite{tao2021function4d}, which comprises 524 high-resolution surfaces. We render 360-degree view images around each surface, which serves as input data. For the quantitative and qualitative evaluation, we use the THuman3.0 dataset~\cite{deepcloth_su2022}, which consists of 20 individuals each captured in various poses (400 models), and the testing set of X-Humans~\cite{shen2023x} (41 models). For each pose, we render 360-degree views of the subject and we apply the reference selection to select the $N_p \times N_r$ references among the images of the subject in different poses, excluding the target one. We evaluate the reconstruction accuracy with three quantitative metrics: the normal reprojection error~\cite{saito2019pifu}, the average point-to-surface Euclidean distance (P2S) and the Chamfer distance (CD), expressed in cm. We further test COSMU with real data collected from various scenarios, including monocular videos (People-Snapshot~\cite{alldieck2018video}, iPER~\cite{lwb2019} datasets) or images of people in different poses and clothing from the Internet, demonstrating its superiority over related works to reconstruct complete 3D human shapes with higher quality details in regions of the body occluded in the input target image (\cref{ssec:comparison_cosmu}). In contrast to prior works, no parametric models are used. We set $N_p=10$ and $N_r=3$. See supplementary for additional implementation details.
\subsection{Ablation Studies}
\label{ssec:ablation}
\begin{figure*}[t!]
\centering
\includegraphics[width=0.9\linewidth]{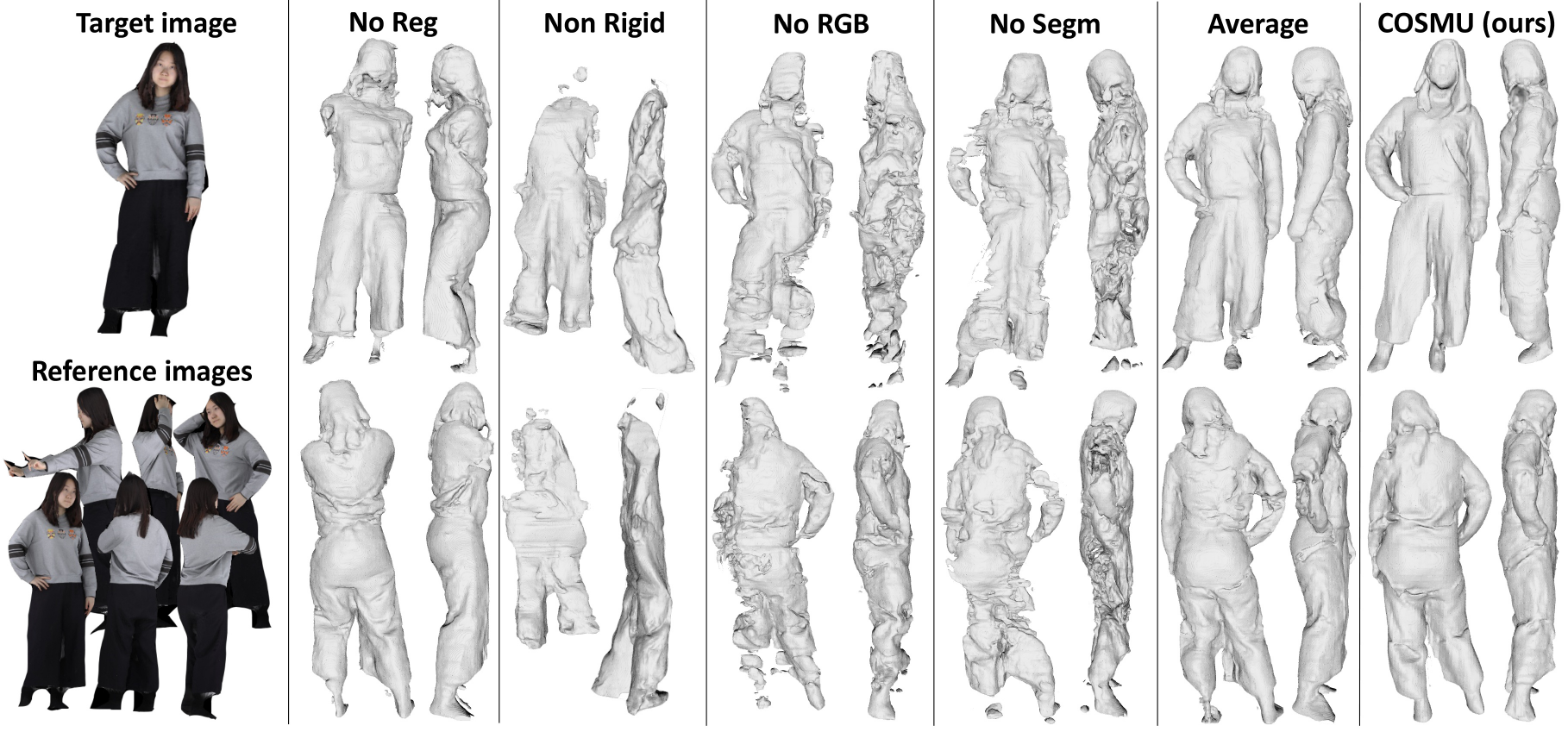}
\caption{Qualitative comparisons between 3D shapes from THuman3.0~\cite{deepcloth_su2022} reconstructed from 2D normal maps generated using different configurations of the body part-based registration algorithm. Front, back and side views of the shapes are shown.}
\label{fig:ablation_shape}
\end{figure*}

\textbf{Energy function for reference selection}:
We conduct an ablation study on the criteria used for selecting the $N_p \times N_r$ references in the body part-based reference selection. The variants explored are: (i) \textbf{Random}: no criteria are applied. $N_p$ references are randomly selected for each group; (ii) \textbf{Only} $\mathbf{b_p}$: only the body part pose similarity is considered; (iii) \textbf{Only} $\mathbf{o}$: the body orientation angle is the only criterion used; (iv) \textbf{Only} $\mathbf{v_p}$: only the body part visibility is considered; (v) $\mathbf{o + v_p}$: the body part pose similarity is excluded; (vi) $\mathbf{o + b_p}$: the body part visibility is omitted; (vii) $\mathbf{v_p + b_p}$: the orientation angle is not used.
\\As illustrated in~\cref{fig:ablation_energy} and supported by the quantitative results of~\cref{tab:ablation_energy}, the 3D shapes obtained with COSMU exhibit fewer artifacts and finer details. The reference selection plays a crucial role in selecting the most suitable references to facilitate the body part-based registration. Omitting any of the criteria in the energy optimization influences the selection of references, leading to suboptimal choices and, consequently,  deteriorating the body part-based registration accuracy and introducing artifacts in the reconstruction.
\begin{table}[!]
\caption{Quantitative evaluation of the application of different energy optimization processes to select reference images from THuman3.0~\cite{deepcloth_su2022}.}
\centering
\resizebox{0.9\linewidth}{!}{\begin{tabular}{c|cccccccc}

\multicolumn{1}{c|}{}       & \multicolumn{1}{c|}{\textbf{Random}} & \multicolumn{1}{c|}{\textbf{Only} $\mathbf{b_p}$} & \multicolumn{1}{c|}{\textbf{Only} $\mathbf{o}$} & \multicolumn{1}{c|}{\textbf{Only} $\mathbf{v_p}$} & \multicolumn{1}{c|}{$\mathbf{o + v_p}$} & \multicolumn{1}{c|}{$\mathbf{o + b_p}$} & \multicolumn{1}{c|}{$\mathbf{v_p + b_p}$}& \multicolumn{1}{c}{\textbf{COSMU}} \\ \hline
\textbf{CD}     & 1.485           & 1.441              & 1.421            & 1.410           & 1.346            &       1.352& 1.326& \textbf{1.304}         \\ 
\textbf{Normal} & .1683           & .1598              & .1641            & .1654           & .1523            & .1529 &  .1503&    \textbf{.1475}          \\ 
\textbf{P2S}    & 1.424           & 1.415              & 1.403            & 1.406           & 1.332            &       1.279& 1.308& \textbf{1.279}        

\end{tabular}

}

\label{tab:ablation_energy}
\end{table}
\\\textbf{Body part-based registration effectiveness}: Directly using $N_r$ selected references for the multi-view reconstruction deteriorates the performance since fusing different poses is challenging in the absence of known correspondences. To avoid this, we introduce a novel body part-based registration method to simulate a multi-view scenario. To prove its effectiveness, we modify the algorithm for registering the $N_p \times N_r$ coarse reference point clouds $\zeta_{ref}$ to the target $\zeta_t$:
\begin{itemize}[noitemsep]
    \item\textbf{No Reg}: To demonstrate the effectiveness of registering the reference point clouds to the target, we omit the body part-based registration step. We adjust the reference selection to select only $N_r$ references by averaging over $p$ the criteria $v^p_g$ and $b^p_g$ in ~\cref{eq:energy}.
    \item\textbf{Non Rigid}: Instead of applying rigid registration among the correspondent body parts of the target and reference point clouds, we apply the state-of-the-art work by Li and Harada~\cite{li2022DeformationPyramid} to perform non-rigid registration between the full-body point clouds. We use $N_r$ references as in \textbf{No reg}.
    \item\textbf{Average}: COSMU registers a target body part with the corresponding body part of a reference selected specifically for that body part. In this configuration, we select only $N_r$ references as in \textbf{No reg} and register all the body parts of the single reference for each group. 
    \item\textbf{No RGB}: The body part classification model $f_{\zeta}$ processes a concatenation of 2D RGB images and segmentation masks. To demonstrate the advantages of using these data, we train and test $f_{\zeta}$ with only segmentation masks.
    \item\textbf{No Segm}: we train and test $f_{\zeta}$ with only RGB images.
\end{itemize}
\begin{table}[t]
\caption{Quantitative evaluation of the application of different configurations of the body part-based registration algorithm on THuman3.0~\cite{deepcloth_su2022}.}
\centering
\resizebox{0.8\linewidth}{!}{\begin{tabular}{c|cccccc}

\multicolumn{1}{c|}{}       & \multicolumn{1}{c|}{\textbf{No Reg}} & \multicolumn{1}{c|}{\textbf{Non rigid}} & \multicolumn{1}{c|}{\textbf{Average}} & \multicolumn{1}{c|}{\textbf{No RGB}} & \multicolumn{1}{c|}{\textbf{No Segm}} & \multicolumn{1}{c}{\textbf{COSMU}} \\ \hline
\textbf{CD}     & 3.546           & 6.932              & 1.314            & 2.249           & 2.189            &       \textbf{1.304}         \\ 
\textbf{Normal} & .2280           & .4832              & .1569            & .2351           & .2114            &      \textbf{.1475}          \\ 
\textbf{P2S}    & 2.820           & 6.024              & 1.309            & 2.169           & 2.005            &       \textbf{1.279}        

\end{tabular}

}

\label{tab:bodypart}
\end{table}
We apply these configurations to generate 2D multi-view normal maps used by the multi-view attention-based implicit model to reconstruct 3D human shapes. Quantitative (\cref{tab:bodypart}) and qualitative results (\cref{fig:ablation_shape}) show that COSMU configuration is the most effective. Multi-view reconstruction requires registered input since pixel-alignment across different views can be incorrect with unregistered views. 
Estimating correspondences between point clouds is challenging and the non-rigid registration cannot produce reliable results. Selecting $N_p$ references for each body part facilitates the body part registration compared to using a single reference. Leveraging RGB and segmentation masks improves the body part classification since the model learns both
semantic and appearance information.
\subsection{Comparisons}
\label{ssec:comparison_cosmu}
\begin{figure*}[!]
\centering
\includegraphics[width=\textwidth,height=0.99\textheight,keepaspectratio]{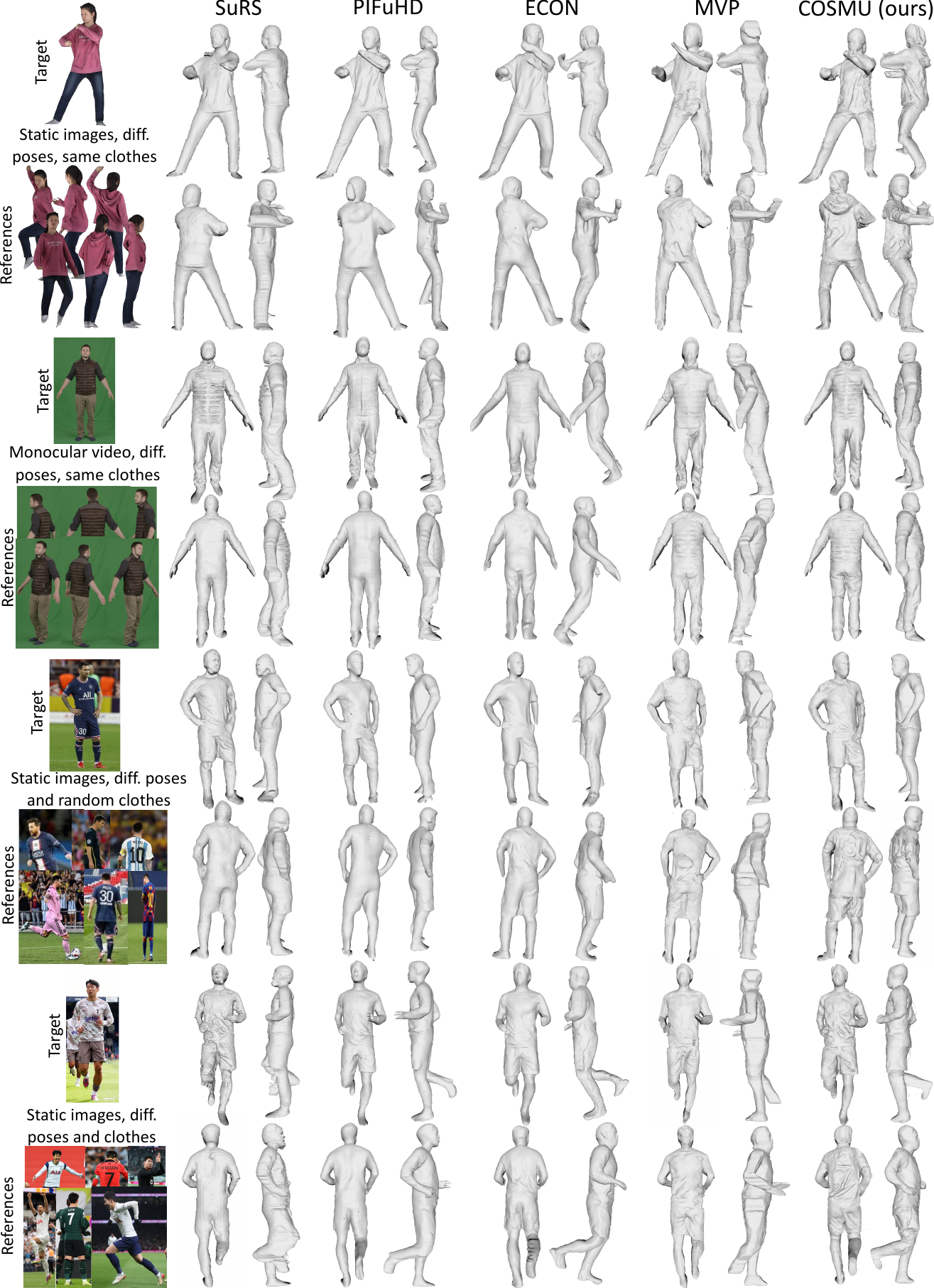}
\caption{Qualitative examples of 3D shapes reconstructed in several scenarios. Static images are from THuman3.0~\cite{deepcloth_su2022} in the first example and from the Internet in the others; monocular videos are from People-Snapshot~\cite{alldieck2018video}. Front, back and side views are shown.}
\label{fig:comparisons}
\end{figure*}

We conduct a comparative evaluation of COSMU against related works that reconstruct complete 3D human shapes from monocular RGB images. We evaluate MVP-Human~\cite{zhu2022mvp} (MVP) and Multi-Pamir~\cite{zheng2021pamir} that reconstruct 3D shapes from unconstrained images leveraging SMPL~\cite{SMPL:2015} in the process. The same $N_r=3$ images selected in the ablation study (\cref{ssec:ablation}) are used as additional views. We evaluate methods that estimate an unseen view of the subject from the target image (PiFU-HD~\cite{saito2020pifuhd}, ICON~\cite{ICONCVPR2022} and ECON~\cite{xiu2023econ}). We then compare COSMU with PiFU~\cite{saito2019pifu} and SuRS~\cite{SuRSECCV2022}, which process a single view without any information on hidden regions. 
To demonstrate the effect of using a lower number of references, we compare COSMU with COSMU-2, which is trained and evaluated by selecting $N_r=1$ reference ($\theta_c=180\degree$). Finally, we evaluate $f_{c}$ (Baseline) that reconstructs the coarse target point clouds $\zeta_t$ without leveraging additional views.  We use the released weights of ECON and PiFU-HD due to the absence of the training code while we train the other approaches with THuman2.0~\cite{tao2021function4d}.
\\\textbf{Quantitative evaluation} is conducted with the 3D shapes reconstructed from single images of THuman3.0~\cite{deepcloth_su2022} and X-Humans~\cite{shen2023x}. The results presented in~\cref{tab:comparisons} demonstrate the superiority of COSMU across all the considered metrics for both datasets.  The best scores overall are highlighted in red while the best scores for approaches that use 2 views are blue. COSMU achieves the highest accuracy while COSMU-2 outperforms approaches that rely on 2 views, demonstrating the benefit of leveraging views from 2D images rather than estimating them from the input. The worst results are obtained with single-view approaches.
\begin{table}[t!]
\caption{Quantitative comparisons with related works for 3D human reconstruction.}
\centering
\resizebox{0.7\linewidth}{!}{\begin{tabular}{c|c|ccc|ccc}

                                       &    \multirow{2}{*}{\textbf{Methods}} &   \multicolumn{3}{c|}{\textbf{THuman3.0}} & \multicolumn{3}{c}{\textbf{X-Humans}} \\ \cline{3-8}    
                                              &  & \multicolumn{1}{c}{CD}    & \multicolumn{1}{c}{Normal} & \multicolumn{1}{c|}{P2S} & \multicolumn{1}{c}{CD}    & \multicolumn{1}{c}{Normal} & \multicolumn{1}{c}{P2S}    \\ \hline
\multicolumn{1}{c|}{\multirow{3}{*}{\rotatebox[origin=c]{90}{\textbf{\textit{1 view}}}}} & \multicolumn{1}{c|}{PiFU~\cite{saito2019pifu}  }      & 1.698          & .1785           & 1.659   & 2.516          & .2591           & 2.492          \\  
\multicolumn{1}{c|}{}                  & \multicolumn{1}{c|}{SuRS~\cite{SuRSECCV2022}    }    & 1.573          & .1693           & 1.554     & 2.406          & .2487           & 2.385          \\ 
\multicolumn{1}{c|}{}                  & \multicolumn{1}{c|}{Baseline}    & 1.566          & .1732           & 1.570 & 2.375          & .2486           & 2.351          \\ \hline
\multicolumn{1}{c|}{\multirow{4}{*}{\rotatebox[origin=c]{90}{\textbf{\textit{2 views}}}}} & \multicolumn{1}{c|}{PiFu-HD~\cite{saito2020pifuhd}  }   & 1.501          & .1663           & 1.512     & 2.139          & .2276           & 2.121          \\ 
\multicolumn{1}{c|}{}                  & \multicolumn{1}{c|}{ICON~\cite{ICONCVPR2022}}        & 1.474          & .1695           & 1.463 & 2.210          & .2395           & 2.208          \\ 
\multicolumn{1}{c|}{}                  & \multicolumn{1}{c|}{ECON~\cite{xiu2023econ}  }      & 1.368          & .1632           & \color[HTML]{3531FF}{1.365}        & 2.153          & .2305           & 2.114         \\ 
\multicolumn{1}{c|}{}                  & \multicolumn{1}{c|}{COSMU-2 }    & \color[HTML]{3531FF}{1.355}          & \color[HTML]{3531FF}{.1593}           & 1.373     & \color[HTML]{3531FF}{2.139}          & \color[HTML]{3531FF}{.2258}           & \color[HTML]{3531FF}{2.104}          \\ \hline
\multicolumn{1}{c|}{\multirow{3}{*}{\rotatebox[origin=c]{90}{\textbf{\textit{4 views}}}}} & \multicolumn{1}{c|}{Multi-PaMIR~\cite{zheng2021pamir}} & 1.401          & .1650           & 1.385    & 2.145          & .2319           & 2.132          \\ 
\multicolumn{1}{c|}{}                  & \multicolumn{1}{c|}{MVP~\cite{zhu2022mvp}}         & 1.348          & .1558           & 1.365    & 2.063          & .2204           & 1.972          \\ 
\multicolumn{1}{c|}{}                  & \multicolumn{1}{c|}{COSMU}      & \color[HTML]{FF0000}{\textbf{1.304}} & \color[HTML]{FF0000}{\textbf{.1475}}  & \color[HTML]{FF0000}{\textbf{1.279}}   & \color[HTML]{FF0000}{\textbf{2.006}} & \color[HTML]{FF0000}{\textbf{.2146}}  & \color[HTML]{FF0000}{\textbf{1.929}}

\end{tabular}}
\label{tab:comparisons}
\end{table}
\textbf{Qualitative results: }\cref{fig:comparisons} illustrates the 3D shapes reconstructed with the considered methods across several scenarios. The first set of shapes is reconstructed by leveraging a collection of images of the same individual in different poses from THuman3.0~\cite{deepcloth_su2022}. 
The target and references of the second set are taken from the same monocular video of a person rotating from People-Snapshot~\cite{alldieck2018video}. 
In the third set, random images of a person in different poses and with a combination of both identical and different clothing are selected from the Internet. 
The fourth set of unconstrained images is created with images of the individual in different poses and clothing. This scenario is the most challenging since there are no images of the person in identical clothing. In all these scenarios, COSMU consistently reproduces higher-quality details in the regions of the 3D shape occluded in the target image, significantly outperforming related works. In addition, the quality of regions visible in the input image is comparable to that of related works. MVP struggles to reproduce details and introduces noise when the subject is wearing different clothing in the references ($3^{rd}$  and $4^{th}$ sets). ECON and PIFu-HD represent details on unseen parts of the model at a significantly lower quality compared to COSMU, which is the only approach that reconstructs the hood on the jumper ($1^{st}$ set). The shapes reconstructed with PIFu-HD and ECON exhibit excessive smoothing in the regions of the human body occluded in the target image. The shapes reconstructed by ECON and MVP are influenced by the incorrect estimation and fitting of the SMPL model, resulting in an unnatural bending of the 3D body. SuRS reproduces high-quality details only in visible regions, with excessive smoothed reconstruction in other regions. See supplementary materials for additional results.
\\\textbf{Limitations}:
COSMU's ability to reconstruct high-quality details is contingent on the availability of reference images. COSMU cannot reproduce high-quality details in a region that is not visible in the reference images. 
Performance deteriorates when the reference images significantly differ from the target. See supplementary for a study on choosing references with varying degrees of similarity to the target. In some examples, COSMU cannot reproduce the same high-quality level of details in small regions of the body, such as the face and hands. The registration algorithm struggles to register small body parts while the multi-view transformer struggles to fuse multiple views of those regions.
\section{Conclusion}
\label{sec:5_conclusion}
We introduce a novel framework to reconstruct complete 3D human shapes from a target RGB image by leveraging a collection of monocular unconstrained images without using parametric models. We first developed a novel multi-view scenario simulation module to generate a set of registered 2D normal maps of the target subject. A multi-view attention-based neural implicit model then estimates an implicit function that represents the subject 3D shape. The evaluation demonstrates that the proposed approach can be applied in various scenarios. Significantly higher quality details are reproduced in non-visible regions of the 3D shape compared to state-of-the-art works. In future works, a solution to enhance the quality of smaller body regions will be investigated and the approach will be extended to reconstruct temporally-coherent dynamic shapes. \\{\footnotesize \textbf{Acknowledgement}: Work supported by BBC Prosperity Partnership AI4ME: Future Personalised Object-Based Media Experiences Delivered at Scale Anywhere EP/V038087.}
\bibliographystyle{splncs04}
\bibliography{main}




\clearpage


%
%
\end{document}